\documentclass[runningheads]{llncs}

\usepackage[T1]{fontenc}
\def\doi#1{\href{https://doi.org/\detokenize{#1}}{\url{https://doi.org/\detokenize{#1}}}}

\usepackage{graphicx}
\usepackage{listings}
\usepackage{amsmath}
\usepackage{amssymb}
\usepackage{xcolor}
\usepackage{caption}
\usepackage[export]{adjustbox}
\usepackage{tabu}
\usepackage{subcaption}
\usepackage[hidelinks]{hyperref}
\usepackage[misc,geometry]{ifsym}

\begin{document}
	
	\title{Anomaly-aware multiple instance learning for rare anemia disorder classification}
	\titlerunning{Anomaly-aware MIL for rare anemia disorder classification}
	
	\author{ 
		Salome Kazeminia \inst{1,2}
		\and
		Ario Sadafi \inst{1,3}
		 \and
		Asya Makhro \inst{4} 
		\and
		Anna Bogdanova  \inst{4}
		 \and
		Shadi Albarqouni  \inst{5,6,7}
		\and
		Carsten Marr \inst{1}$^{(\textrm{\Letter})}$
	} 
		
	\institute{
		Institute of AI for Health, Helmholtz Munich – German Research Center for Environmental Health, Neuherberg, Germany
	 	\and
	 	Technical University of Munich, Munich, Germany
	 	\and
	 	Computer Aided Medical Procedures, Technical University of Munich, Munich, Germany
	 	\and
	 	Red Blood Cell Research Group, Institute of Veterinary Physiology, Vetsuisse Faculty and the Zurich Center for Integrative Human Physiology, University of Zurich, Zurich, Switzerland
	    \and
	    Clinic for Diagnostic and Interventional Radiology, University Hospital Bonn, Germany
	    \and
	    Helmholtz AI,  Helmholtz Munich – German Research Center for Environmental Health, Neuherberg, Germany
	    \and
	    Faculty of Informatics, Technical University Munich, Munich, Germany
}
\authorrunning{S. Kazeminia et al.}
\maketitle 

\begin{abstract}
	Deep learning-based classification of rare anemia disorders is challenged by the lack of training data and instance-level annotations. 
	Multiple Instance Learning (MIL) has shown to be an effective solution, yet it suffers from low accuracy and limited explainability. 
	Although the inclusion of attention mechanisms has addressed these issues, their effectiveness highly depends on the amount and diversity of cells in the training samples.
	Consequently, the poor machine learning performance on rare anemia disorder classification from blood samples remains unresolved.
	In this paper, we propose an interpretable pooling method for MIL to address these limitations. 
	By benefiting from instance-level information of negative bags (i.e., homogeneous benign cells from healthy individuals), our approach increases the contribution of anomalous instances.  
	We show that our strategy outperforms standard MIL classification algorithms and provides a meaningful explanation behind its decisions. 
	Moreover, it can denote anomalous instances of rare blood diseases that are not seen during the training phase.
	
	\keywords{Multiple instance learning \and Anomaly pooling \and Rare anemia disorder \and Red blood cells}
	
\end{abstract}

\section{Introduction}
The appearance of human red blood cells changes directly with their volume. 
Typically red blood cells have a discoid concave shape, and as their volume increases or decrease in physiological conditions, they shrivel into star-shaped cells or swell into spherical shapes~\cite{huisjes2020density}. 
In hereditary hemolytic anemias (HHAs), a cell's membrane is not formed properly and cannot accommodate volume changes, resulting in the formation of anomalous morphologies.
Membrane instability and reduced deformability are other factors in a number of HHAs, such as hereditary spherocytosis, sickle cell disease, and thalassemia, where improper membrane formation leads to irregularly formed cells.

Tools for close monitoring of blood samples in HHA patients is crucial for diagnosis~\cite{huisjes2018digital}, disease progression~\cite{fermo2021screening}, and severity estimation~\cite{sadafi2021sickle}. 
However, since most cells in a patient's sample have no evident morphological feature relevant to the HHA, identifying hallmark cells is a needle-in-the-hay-stack search.
Furthermore, presence of a few anomalous cells in a sample does not necessarily link to an underlying condition making the diagnosis even more challenging. 

In addition, proper single cell annotation for supervised model training is a cumbersome, expensive task and introduces intra-expert variablity.
Thus, the automated analysis of blood samples is a perfect problem for multiple instance learning-based approaches. 
Multiple  Instance Learning (MIL) is a weakly supervised learning method, where only bag-level labels are used~\cite{campanella2019clinical}.
Notably, a bag with a negative label does not contain any positive instance, while bags with a positive label may contain negative instances.
In other words, the class of a bag depends on the existence of one or more class-related instance.
Thus, the learning signal, back-propagated to the instance-level feature extractor may be very weak and ineffective.
To address this problem, MIL-based deep classifiers require a large number of training data to perform promisingly~\cite{campanella2019clinical}.
Providing such data is particularly difficult in the case of rare disease.
A possible solution is the deployment of intelligent pooling methods to magnify the learning signal back-propagated from informative instances~\cite{lu2021data}.
In basic MIL classifiers, when the bag contains a small number of class-related instances, a typical max or average pooling cannot to provide such magnification~\cite{lu2021data}. 

Another requirement for health AI applications is explainability.
MIL classifiers should thus be able to specify HHA-related structures of instances anomalies.
Recently, attention-based pooling mechanisms showed a good performance in tackling both learning signal magnification and explainability challenges~\cite{ilse2018attention,lu2021data,sadafi2020attention}. 
However, attention mechanisms are prone to fail in scenarios where only a few training samples are available, and consequently, the algorithm can not identify relevant instances in the bags correctly.
In HHA classification, such a problem leads to noisy attention scores where lots of disorder-relevant (positive) cells receive low attention, and the distribution of attention on non-disorder (negative) cells would be non-uniform. 
Here, a strategy accounting for the diversity of disorder-relevant cells and data imbalance is required. 

This paper introduces an anomaly-aware pooling method, which is able to address limitations of the attention mechanism with an efficient amount of training data.
While instance-level information for positive bags is not available, we exploit negative bag labels that apply to all instances of the bag and estimate the distribution of negative instances using Bayesian Gaussian mixture models. 
By measuring the distance of instances to this distribution, we define an anomaly score that is used for pooling in addition to learned attention scores.
Our method's performance is explainable by specifying the contribution of instances in the final bag classification.
Also, our proposed anomaly score can specify anomalous instances that are rare and haven't been seen during the training. 
We make our code publicly available at \url{https://github.com/marrlab/Anomaly-aware-MIL} .

\section{Method}
\begin{figure}[t]
	\includegraphics[width=\textwidth]{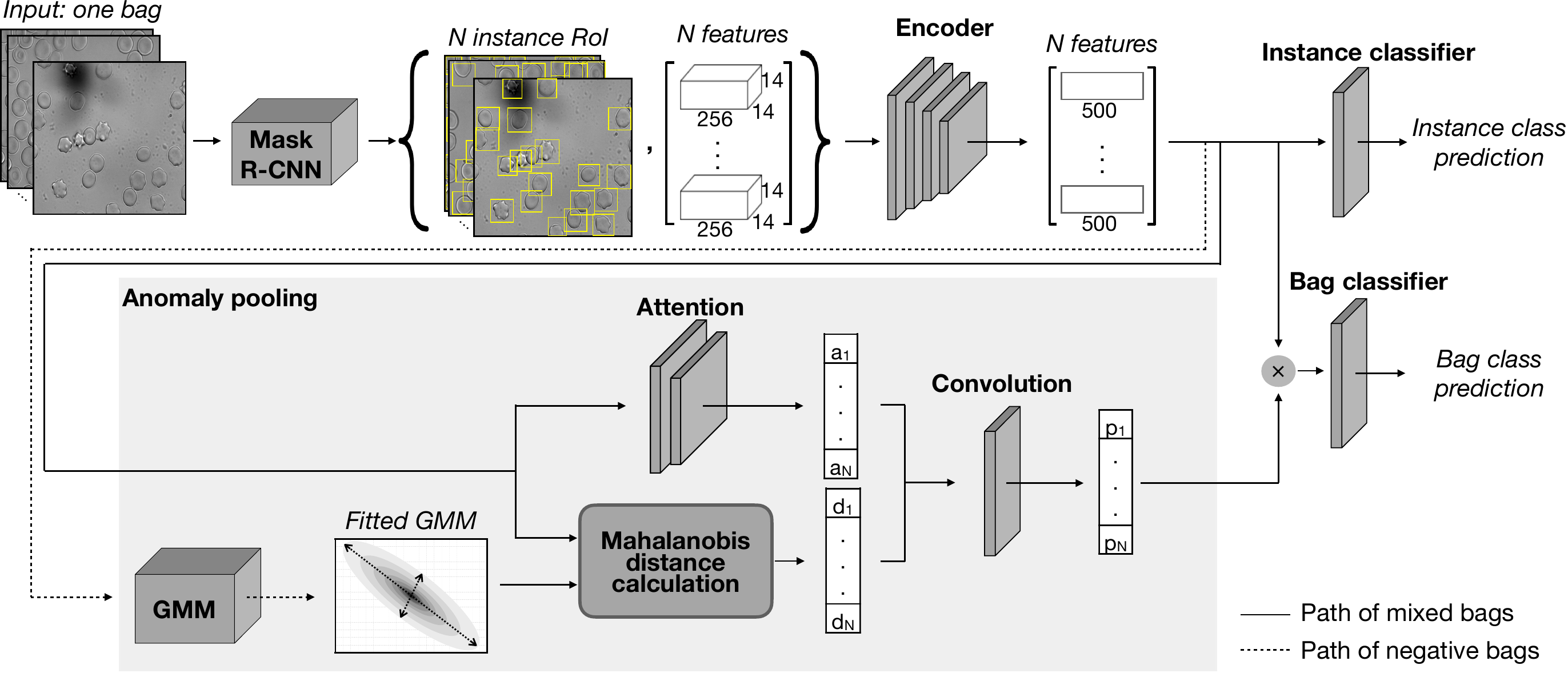}
	\caption{Overview of our proposed anomaly-aware MIL classifier. A Mask R-CNN outputs detected instance bounding boxes and basic instance features. Then, an encoder maps those features to a latent representation. Dashed path: A GMM distribution is fitted to instances from negative bags $B_{neg}$, and its parameters are saved. Solid path: Attention scores $a_{n\in N}$ and anomaly scores $d_{n\in N}$ are estimated and combined through a convolution layer to generate pooling weights $p_{n\in N}$. Finally, a fully connected layer predicts the bag label.}
	\label{fig:diagram}
\end{figure}

Our MIL classifier $f_{\theta}$ consists of an instance detector, a deep instance feature extractor $h_{\psi}$, and an attention-anomaly pooling mechanism that assigns a contribution weight to each instance. 
The input of our classifier is a bag of instances $\{I_{1}, ..., I_{N}\} \in B$; the output is a class $c_{i}$ from possible classes defined in $C$.
In our target application, bags are red blood cells from brieghtfield microscopic images of blood samples, each containing between 12 to 45 cells (See Fig. \ref{fig:diagram}).

\subsection{Instance-level feature extraction}
\subsubsection{Preprocessing.} 
Since instance-level information plays a crucial role in bag classification and interpretation, we need to detect them and follow their footprint in the final classification result.
For instance detection, we prefer two-stage object detection methods due to the higher accuracy of their performance in comparison with one-stage methods~\cite{fujita2020cell}.
As a preprocessing step, any two-stage object detector can be used.
We employ Mask R-CNN~\cite{he2017mask} with ResNet~\cite{he2016deep} backbone since it provides an accuracy above $91\%$ in detecting instances in microscopic images~\cite{fujita2020cell,bessis2012corpuscles}.
In our setting, the instance detector takes the bag $B$ and outputs a bounding box and a feature map for each instance, regardless of their class.
We use these feature maps as instances $I_{n} \in \{I_{1}, ..., I_{N}\} = B$.
Mask R-CNN extracts features of each instance independently, regardless of its size, scale, and location. 
This simple step is extremely beneficial in our MIL training as it avoids challenges stemming from non-uniform information distributions of instances, which is widely researched in the literature~\cite{li2019multi,shi2020loss,bi2021local}. 

\subsubsection{Instance representation.} 
Using a 4-layer convolutional encoder $h_{\psi}: {\mathbb{R}}^{m} \rightarrow {\mathbb{R}}^{k}$, we map each instance $I_{n}$ to a point $z_{n}$ in the $k$-dimensional representation space $Z$:
\begin{equation}
	\label{encoder}
	z_{n} = h_{\psi}(I_{n}),	\quad n \in N
\end{equation}

\subsection{Anomaly-aware pooling}
\subsubsection{Negative distribution estimation.}
We estimate the distribution of negative instances in our latent space and detect anomalies by measuring their distance to this distribution.
As a prior assumption, we consider a Gaussian Mixture Model (GMM) probability distribution on negative instance representations.
Due to Bayes' rule, the posterior probability distribution would be a GMM as well.
Therefore we fit a GMM model $g_{\Gamma}$ on negative instance representations.
To estimate $\Gamma$, we use an Expectation-Maximization (EM) algorithm that miximizes the likelihood $p(\Gamma|Z_{neg})$ given $Z_{neg}:\{z_{j}|I_{j} \in B_{neg}\}$
\begin{equation}
	\label{GMM}
	p(\Gamma|Z_{neg})= \sum_{i} \tilde{\gamma_{i}} \mathcal{N}(\tilde{\mu_{i}},\tilde{\Sigma_{i}}),
\end{equation}
where $\tilde{\gamma}$, $\tilde{\mu_{i}}$, and $\tilde{\Sigma_{i}}$ are estimated weight, mean, and covariance matrix of a Gaussian distribution, characterizing the $i^{th}$ GMM component, respectively.
When the algorithm converges, we save the model's parameters as a reference for anomaly scoring.

\subsubsection{Anomaly scoring.} 
For anomaly scoring we measure the Mahalanobis distance between instances and the $g_{\Gamma}$
\begin{equation}
	\label{Mahalanobis}
	d_{n} =  \sqrt{(z_{n}-\mu)^{T} \Sigma^{-1} (z_{n} - \mu)},
\end{equation}
where $\mu$ and $\Sigma$ are mean and covariance of the $g_{\gamma}$ distribution.
We have chosen Mahalanobis distance over other distances since it considers the distribution's covariance in its distance measurement.

\subsubsection{Attention.}
The attention mechanism is a method to extract meaningful instances contributing to the final decision of the model.
Specially, in MIL it showed a good performance in both classification and interpretation of deep networks~\cite{ilse2018attention,sadafi2020attention,lu2021ai,lu2021data,wu2021combining}.
Our deep attention network $v_{\omega}$ estimates an attention score $a_{n}$ for each instance $I_{n}$ as
\begin{equation}
	\label{attention}
	a_{n} = v_{\omega}(z_{n}).
\end{equation}

\subsubsection{Pooling.}
Our pooling mechanism combines attention $a_{n}$ and anomaly $d_{n}$ scores to estimate a contribution weight for each instance and conclude the latent representation of the bag:
\begin{equation}
	\label{pooling}
	z_{B} = \sum_{n=0}^{N} (W_{D_{n}}d_{n} + W_{A_{n}}a_{z_{n}})z_{n}.
\end{equation}
Here $W_{D}$ and $W_{A}$ are respectively learned parameters, to specify the intervention share of attention and anomaly scores. 
Finally, a fully connected layer $f_{\theta}$ maps the $k$-dimensional representation $z_{B}$ to a class $c_{i}$ (see Fig. \ref{fig:diagram}).
Both anomaly and attention scores of am instance are independently calculated and fed to a 1×1 convolution to estimate the final pooling weight of the corresponding instance. 
This makes our pooling method permutation invariant.

\subsection{Optimization}
The functionality of $v_{\omega}$ and fitting $g_{\Gamma}$ on negative instances highly depends on the encoded instances by $h_{\psi}$.
To boost the training procedure, we start with a single instance classifier (SIC) $s_{\zeta}$ using noisy labels for learning~\cite{sadafi2020attention}.
In other words, we assign the bag label to all of its instances and calculate a cross-entropy loss ($\mathrm{CE_{SIC}}$) to optimize the model in the first epochs. 
Then, by increasing the epoch number $e$, we gradually decrease SIC's impact on the training and conversely increase the MIL cross-entropy loss ($\mathrm{CE_{MIL}}$) contribution.
Therefore, the final loss function, considering a dynamic coefficients $\beta(e)$, is defined as:
\begin{equation}
	\label{loss}
	L(\theta,\zeta,\omega,\psi) = (1-\beta(e))\mathrm{CE_{MIL}} + \beta(e) \mathrm{CE_{SIC}},
\end{equation}

\section{Experiment}
We applied our classifier on microscopic images of blood samples from HHA patients.
In particular, we evaluated three aspects of our method's functionality: 1) classification, 2) explainability, and 3) rare anomaly recognition.

\subsubsection{Dataset.}
Patients previously diagnosed with hereditary spherocytosis, xerocytosis, thalassemia, and sickle cell disease were enrolled in the CoMMiTMenT study. 
The study protocols were approved by the Medical Ethical Research Board of the University Medical Center Utrecht, the Netherlands, under reference code 15/426M and by the Ethical Committee of Clinical Investigations of Hospital Clinic, Spain (IDIBAPS) under reference code 2013/8436. 
Further blood samples of SCD patients were produced for MemSID (NCT02615847) clinical trial. 
The study protocol for it was approved by the Ethics committee of Canton Zurich (KEK-ZH 2015-0297) and the regulatory authority. 
Our dataset contains 3630 microscopic images taken from 71 patients/healthy individuals at different times and under variant treatments. 
The distribution of data in different classes is as follows: Sickle Cell Disease (SCD, 13 patients, 170 image sets), Thalassemia (Thal, 3 patients, 25 image sets), Hereditary Xerocytosis (Xero, 9 patients, 56 image sets), and Hereditary Spherocytosis (HS, 13 patients, 89 image sets). 
Also, we have a healthy control group (33 individuals, 181 image sets), used as negative bags. 

\subsubsection{Training.}
For training we considered $3-$fold cross-validation and 100 epochs.
We have two training sets: i) one set containing only control samples to estimate GMM, and ii) an other set containing a mixture of samples from all classes to train our network.
We considered one GMM component with full covariance and iterated an EM algorithm up to 100 times.
Mixed data are fed to the MIL classifier and every $5$ epoch, control set is fed to the model to calibrate GMM parameters with updates.
For optimization, we use the Adam optimizer~\cite{reddi2019convergence} with the learning rate of $5\times 10^{-5}$ and weight decay of $10^{5}$.
We chose the dynamic SIC loss coefficient $\beta(e)$ as $0.95^{e}$. 

\subsection{Classification}
\begin{figure}[]
	\centering
	\includegraphics[width=\textwidth]{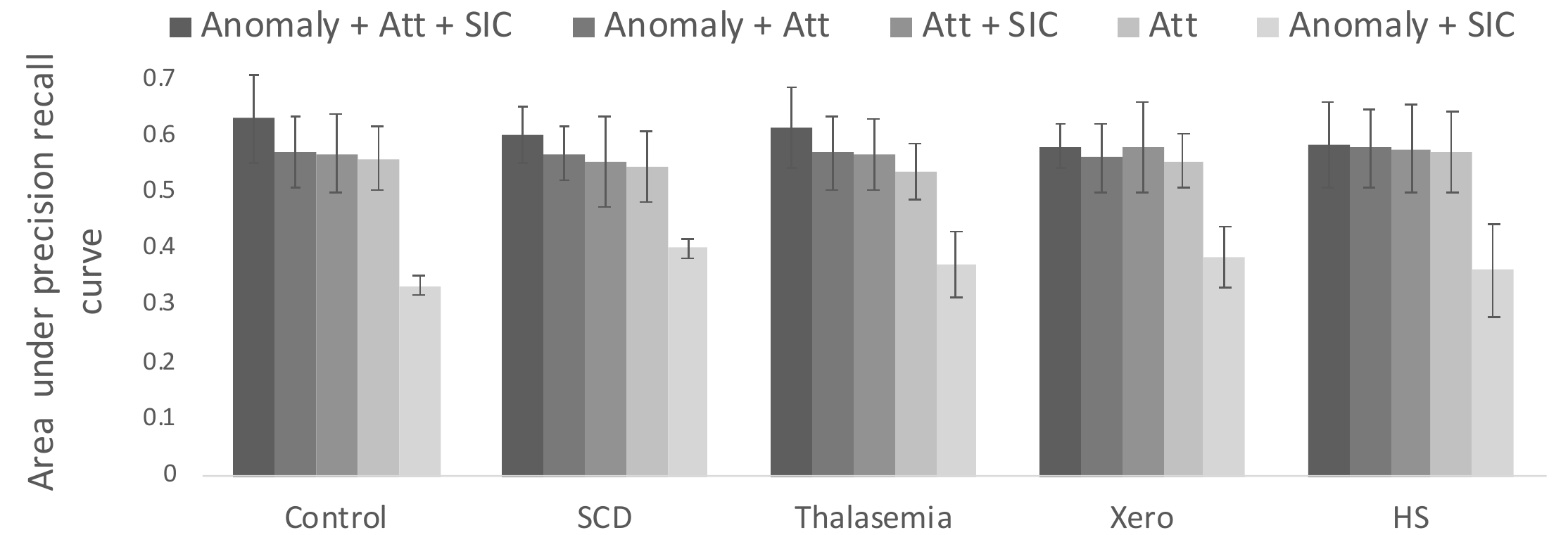}
	\caption{Average and standard deviation of the class-wise area under precision recall curve in five runs. In all classes, our novel method outperforms others.}
	\label{fig:classification}
\end{figure}

Our metrics for classification evaluation are average accuracy, F1 score, and area under the precision recall curve (AUC).
Table \ref{classification} shows our results (Anomaly + Att + SIC) in comparison with other strategies proposed by Ilse et al.~\cite{ilse2018attention} (Att) and Sadafi et al.~\cite{sadafi2020attention} (Att $+$ SIC).
In addition, as an ablation study, we show the performance of our proposed anomaly scoring without SIC loss (Anomaly + Att) and attention mechanism (Anomaly).
The class-wise performance of our proposed method compared to other methods is shown in Fig. \ref{fig:classification}.
\begin{table}[t]
	\centering
	\caption{Average classification performance of our proposed method compared to four baselines. The numbers show the mean value along with the standard deviation of metrics measured in 5 runs.}
	\label{classification}
	\begin{tabular}{l|l|l|l}
		\hline
		\textbf{Method} & \textbf{Accuracy} & \textbf{F1-score} & \textbf{AU RoC} \\
		\hline
		Att~\cite{ilse2018attention} & 0.752$\pm$0.009 & 0.741$\pm$0.010 & 0.826$\pm$0.009 \\
		Att + SIC~\cite{sadafi2020attention} & 0.770$\pm$0.008 & 0.757$\pm$0.009 & 0.808$\pm$0.013 \\
		Anomaly & 0.630$\pm$0.029 & 0.604$\pm$0.026 & 0.648$\pm$0.025\\
		Anomaly + Att & 0.760$\pm$0.019 & 0.743$\pm$0.021 & 0.833$\pm$0.006 \\
		Anomaly + Att + SIC & \textbf{0.787$\pm$0.015} & \textbf{0.770$\pm$0.016} & \textbf{0.854$\pm$0.008} \\
		\hline
	\end{tabular}
\end{table}

\subsection{Explainability}
\begin{figure}[t]
	\includegraphics[width=\textwidth]{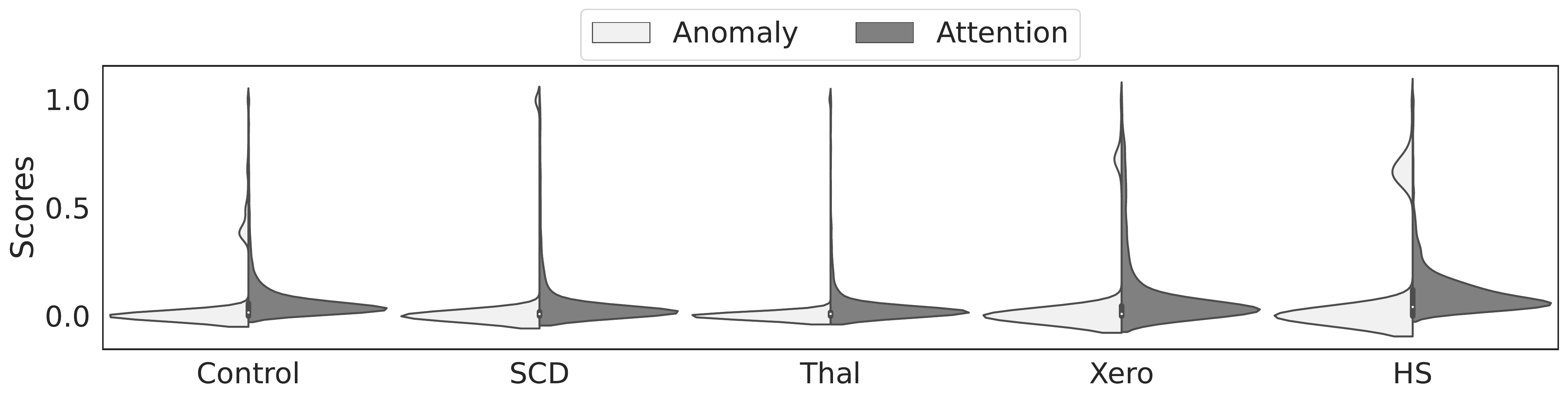}
	\caption{Distribution of anomaly scores compared to attention scores in every class. Our anomaly scoring mechanism can distinguish a higher fraction of disorder-relevant cells.}
	\label{chart_interpretation}
\end{figure}

Our method's anomaly scoring mechanism is interpretable at the instance level.
In Fig. \ref{fig:interpretation}, for each class, we visualized a sample image, in which anomaly and attention scores are depicted by the color of the cell's bounding box.
The anomaly mechanism can distinguish class-related cells better than the attention mechanism, which is particularly striking for the SCD case, where irregular elliptoid cells are missed by the attention mechanism (Fig. \ref{fig:interpretation}).
In Fig. \ref{chart_interpretation} we show the distribution of scores assigned to cells in the test data.
The frequency of both anomaly and attention scores are focused around zero, which is compatible with a majority of healthy cells (see also Fig. \ref{chart_interpretation}).
On the other hand, for HHA, a small distribution of high scores notifies the anomalous cells.
These cells show HHA properties.
\newcommand\w{0.175}
\newcommand\wb{0.024}
\begin{figure}[t]
	\centering
	\begin{subfigure}[b]{0.02\textwidth}
		\centering
		\makebox[5pt]{\raisebox{27pt}{\rotatebox[origin=c]{90}{Anomaly}}}%
	\end{subfigure}
	\hfill
	\begin{subfigure}[b]{\w\textwidth}
		\centering
		\includegraphics[width=\textwidth]{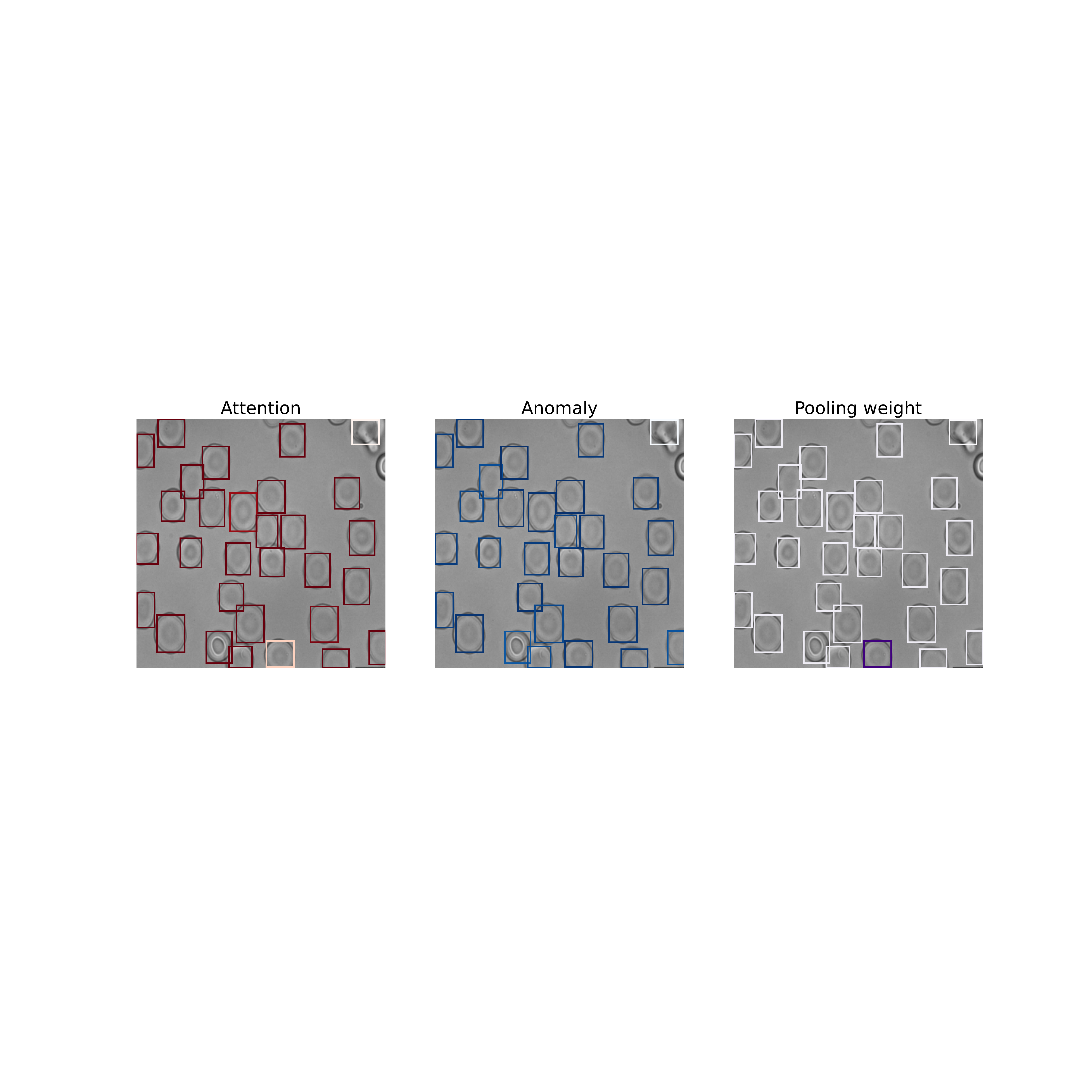}
	\end{subfigure}
	\begin{subfigure}[b]{\w\textwidth}
		\centering
		\includegraphics[width=\textwidth]{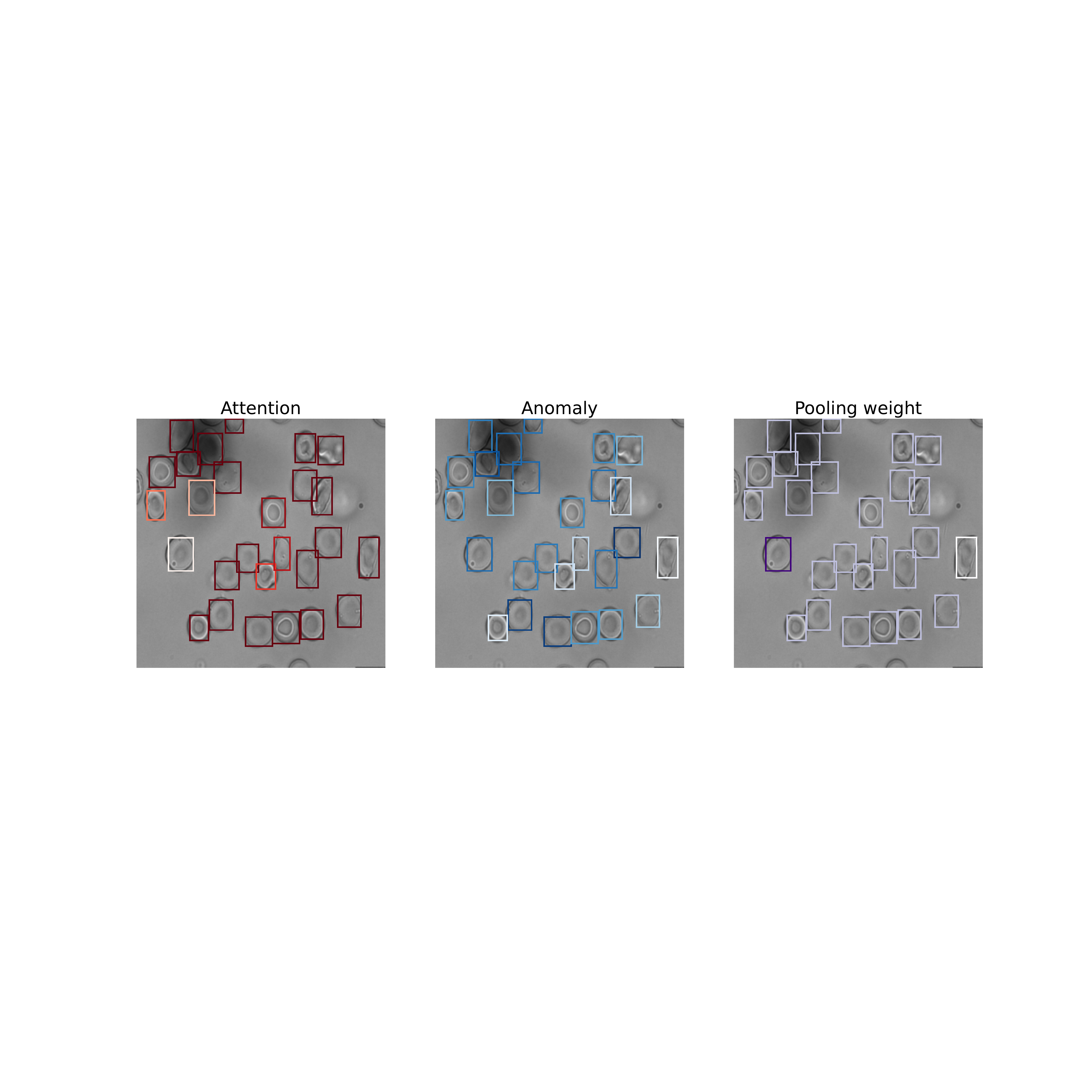}
	\end{subfigure}
	\begin{subfigure}[b]{\w\textwidth}
		\centering
		\includegraphics[width=\textwidth]{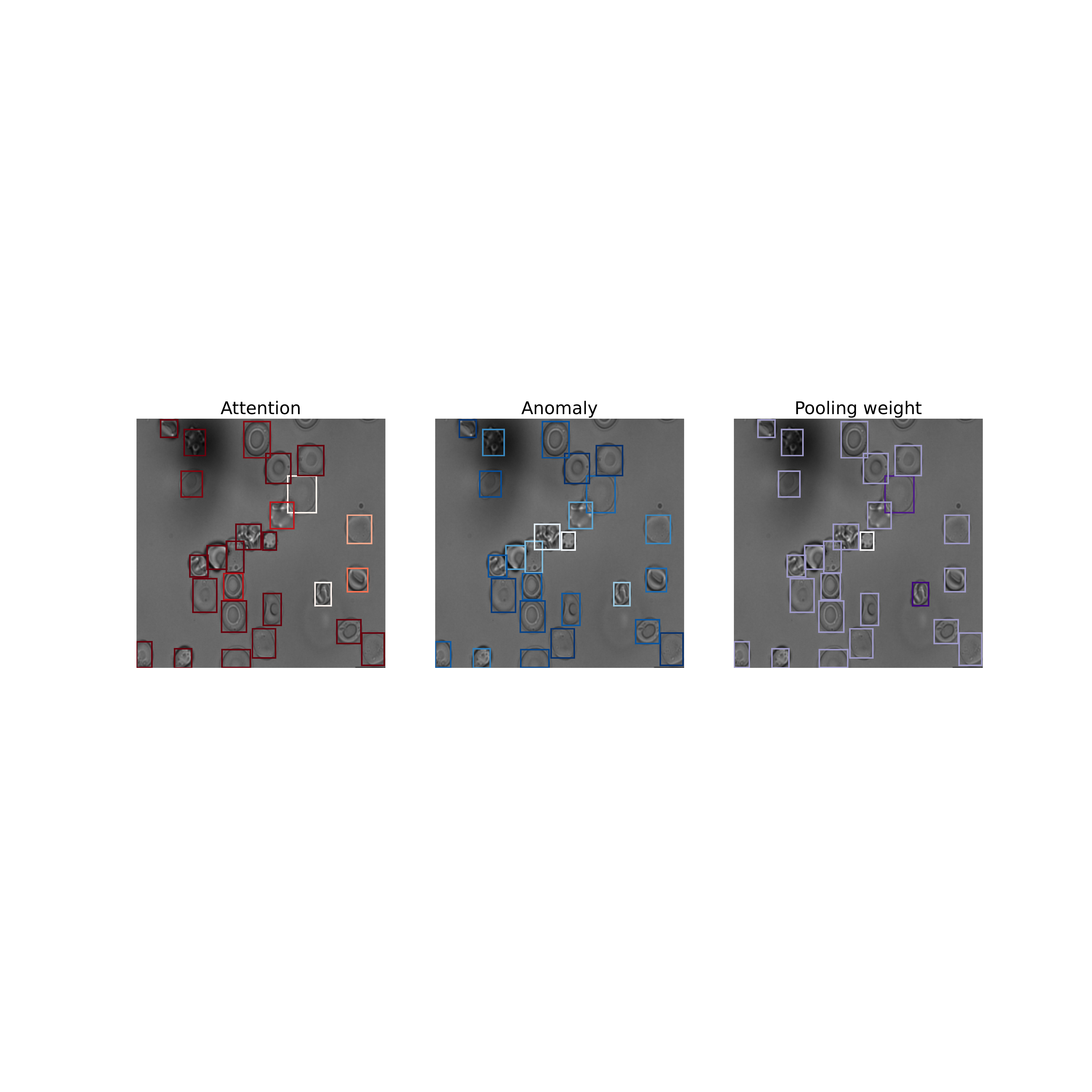}
	\end{subfigure}
	\begin{subfigure}[b]{\w\textwidth}
		\centering
		\includegraphics[width=\textwidth]{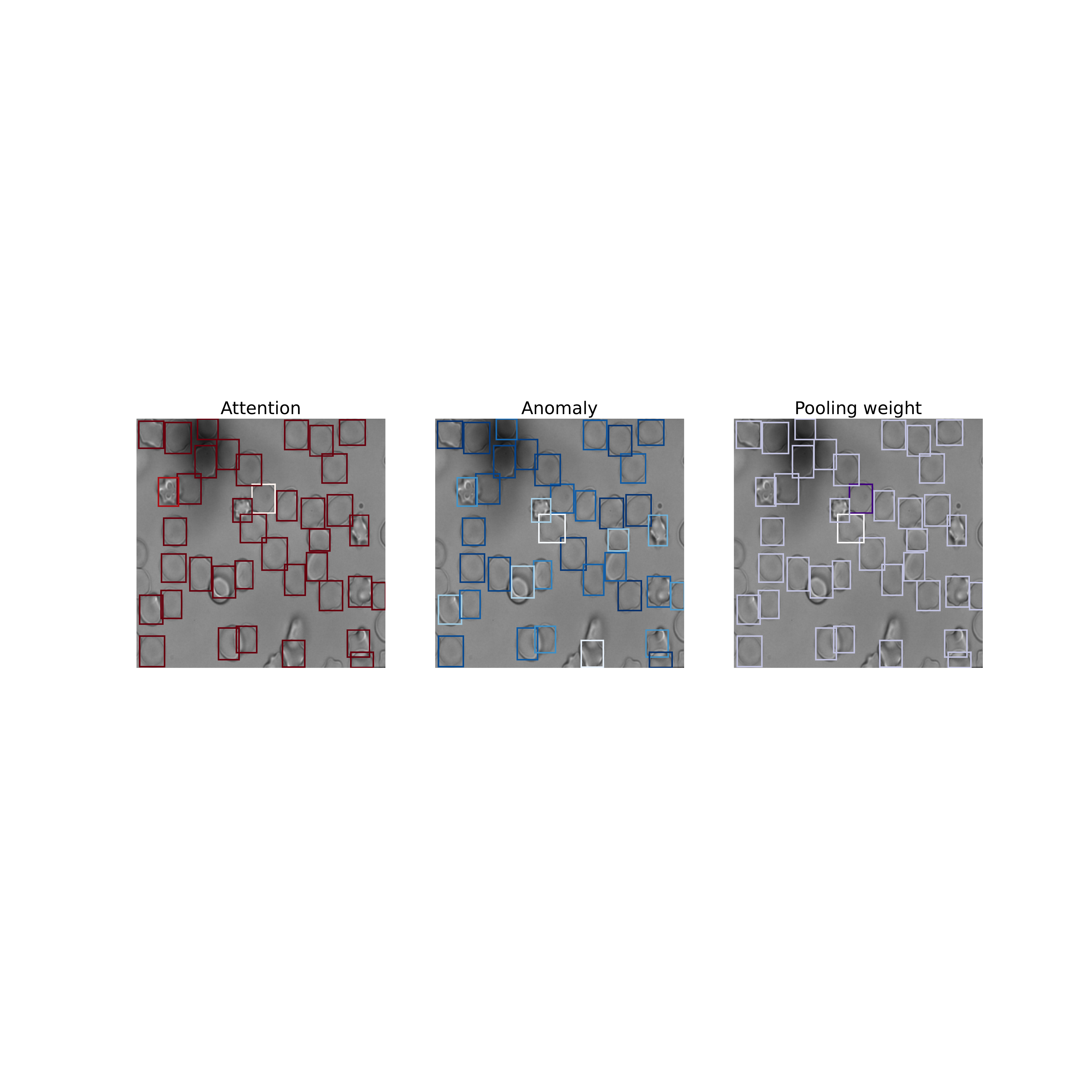}
	\end{subfigure}
	\begin{subfigure}[b]{\w\textwidth}
		\centering
		\includegraphics[width=\textwidth]{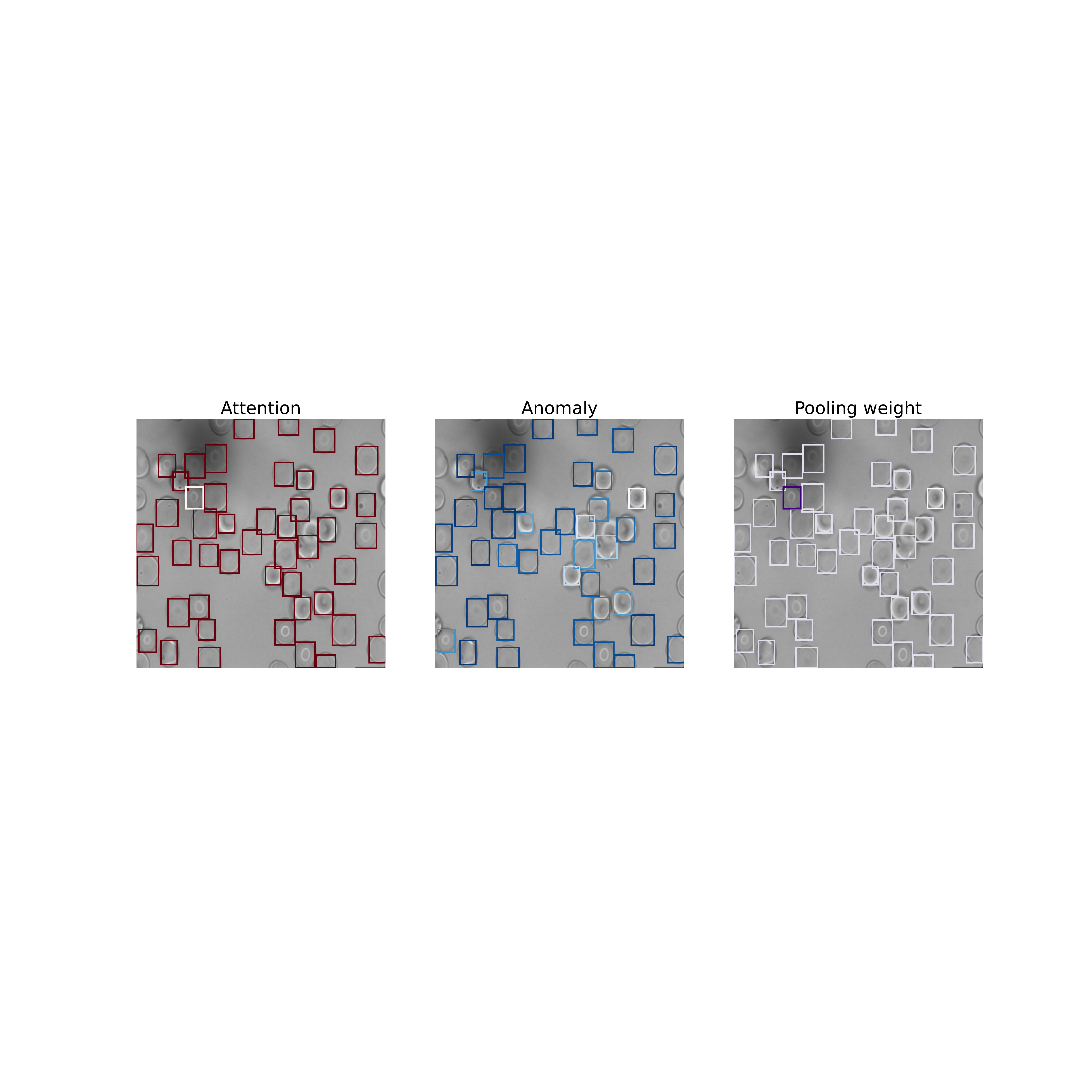}
	\end{subfigure}
	\begin{subfigure}[b]{\wb\textwidth}
		\centering
		\includegraphics[width=\textwidth]{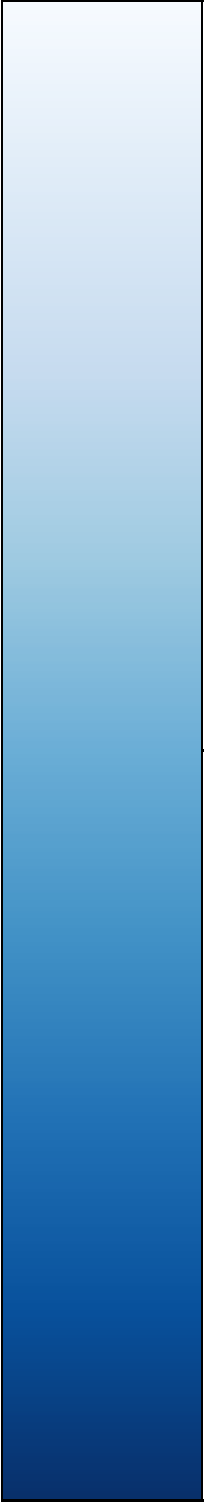}
	\end{subfigure}
	\begin{subfigure}[b]{0.0142\textwidth}
		\centering
		\includegraphics[width=\textwidth]{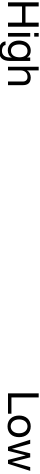}
	\end{subfigure}
	\\
	\begin{subfigure}[b]{0.02\textwidth}
		\centering
		\makebox[5pt]{\raisebox{44pt}{\rotatebox[origin=c]{90}{Attention}}}%
	\end{subfigure}
	\hfill
	\begin{subfigure}[b]{\w\textwidth}
		\centering
		\includegraphics[width=\textwidth]{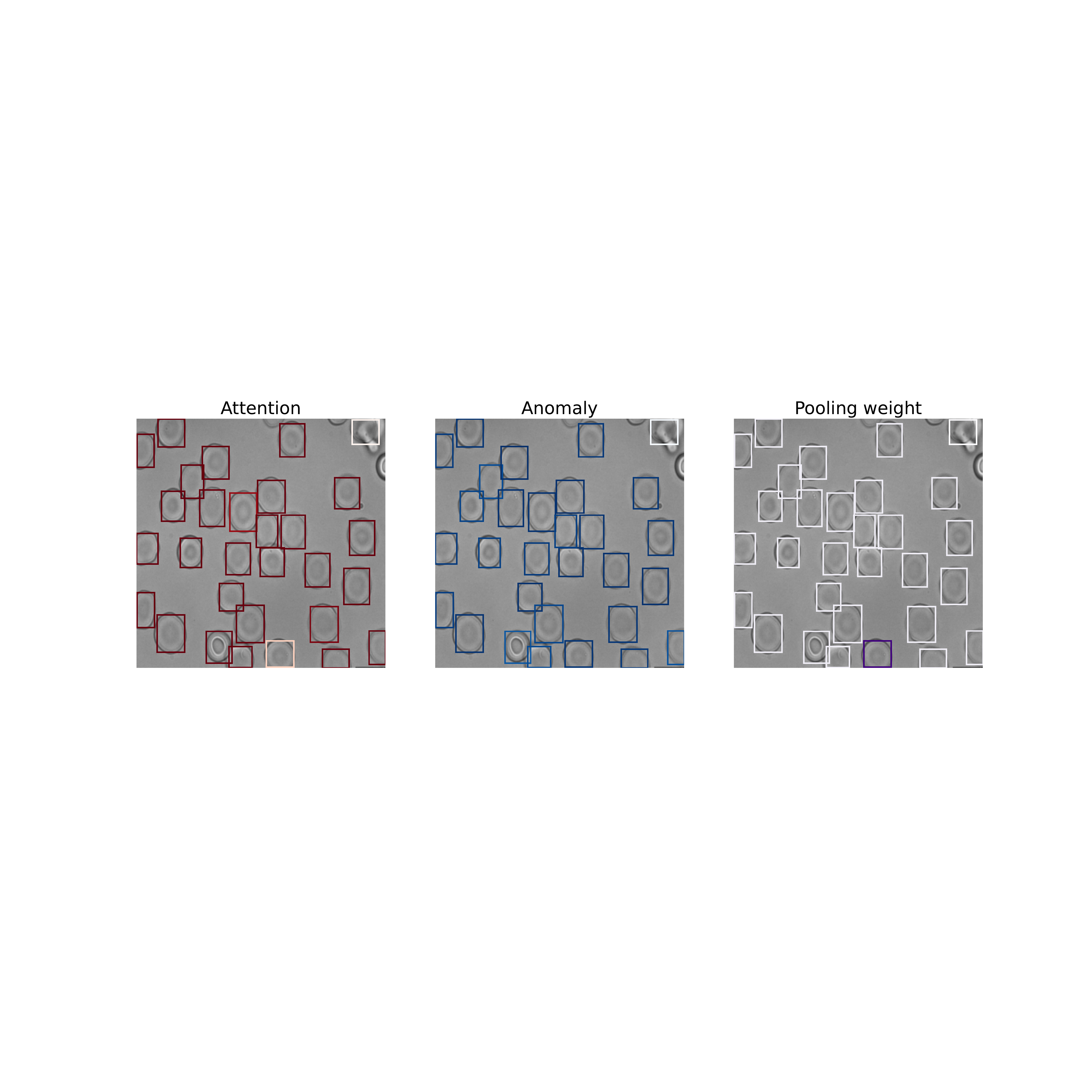}
		\caption*{Control}
	\end{subfigure}
	\begin{subfigure}[b]{\w\textwidth}
		\centering
		\includegraphics[width=\textwidth]{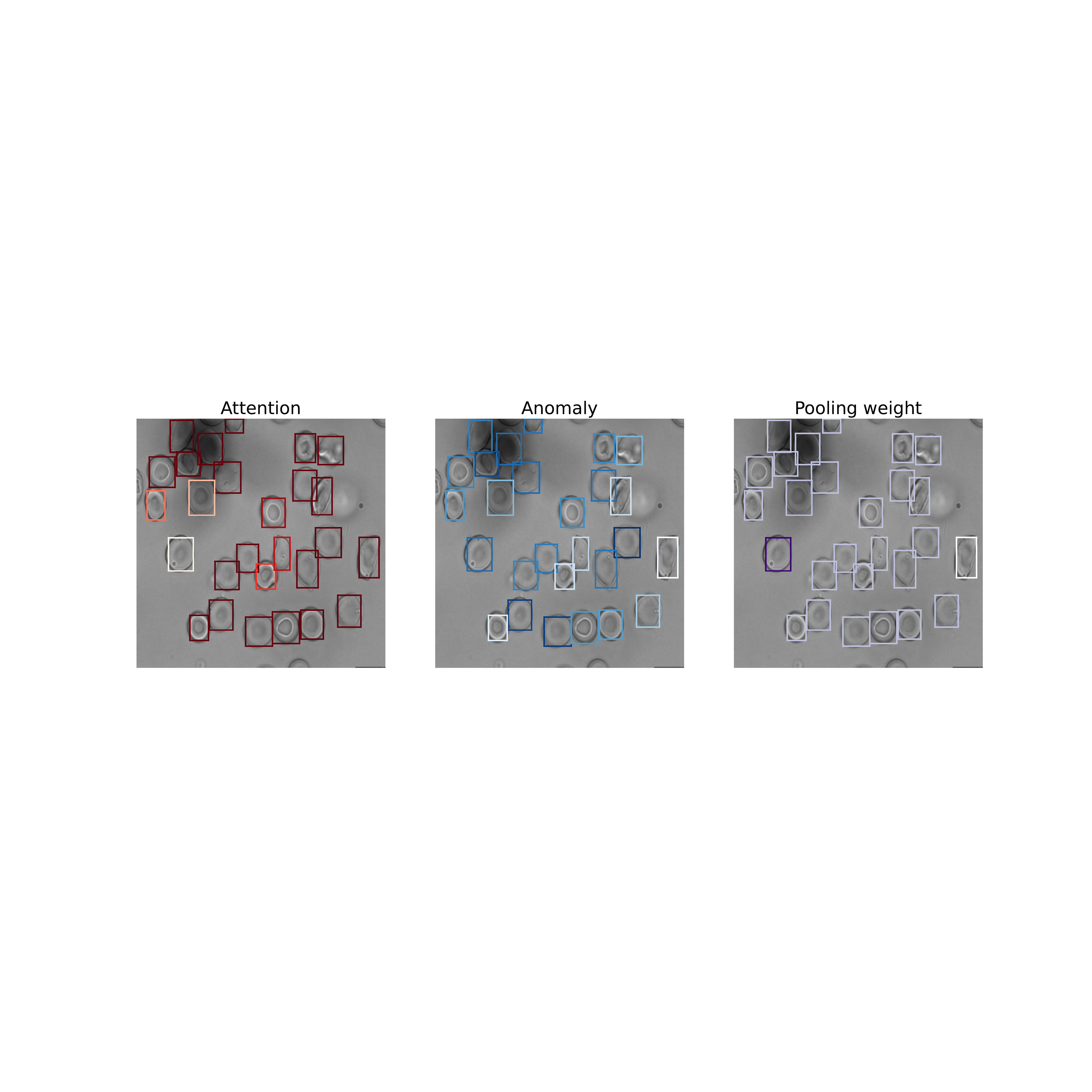}
		\caption*{SCD}
	\end{subfigure}
	\begin{subfigure}[b]{\w\textwidth}
		\centering
		\includegraphics[width=\textwidth]{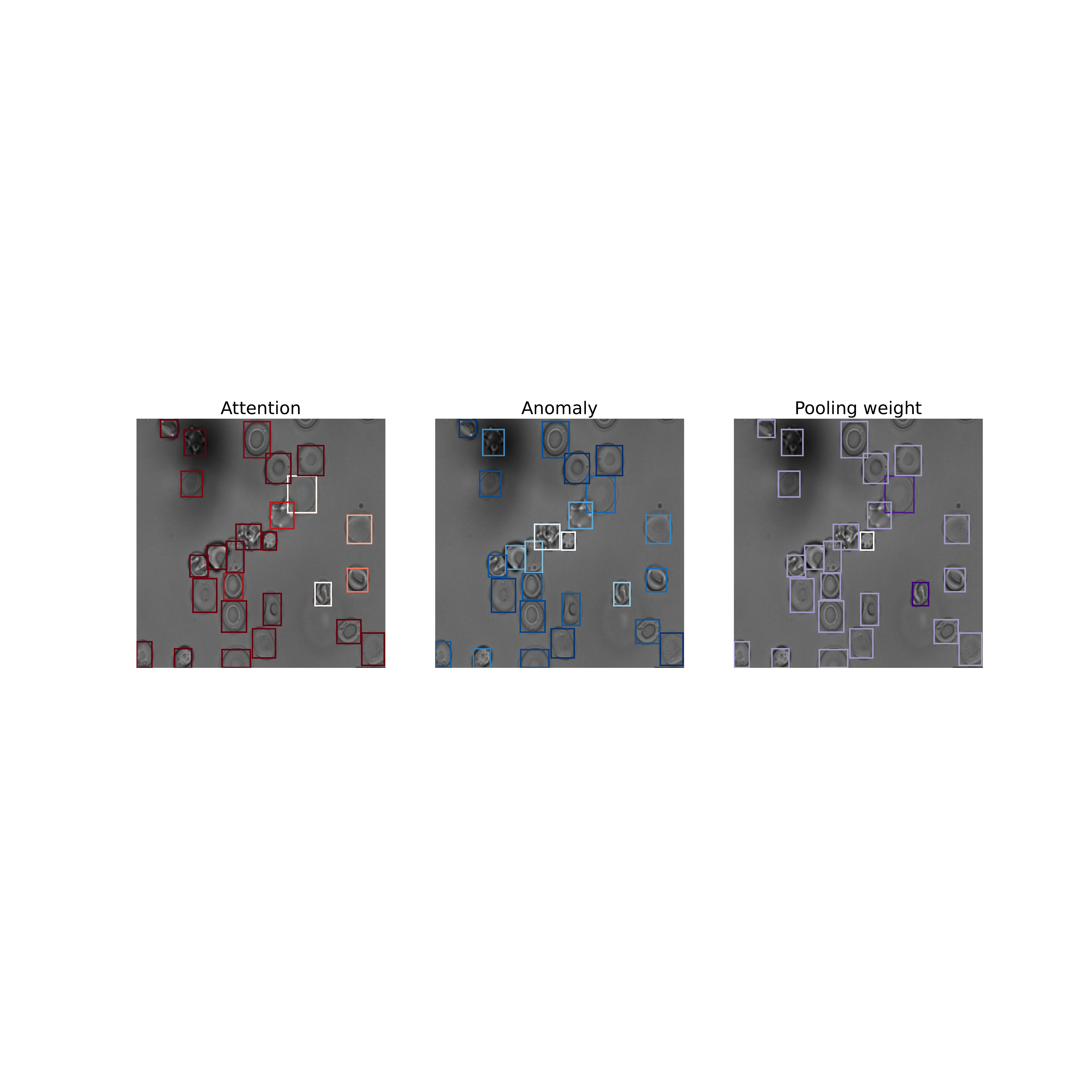}
		\caption*{Thal}
	\end{subfigure}
	\begin{subfigure}[b]{\w\textwidth}
		\centering
		\includegraphics[width=\textwidth]{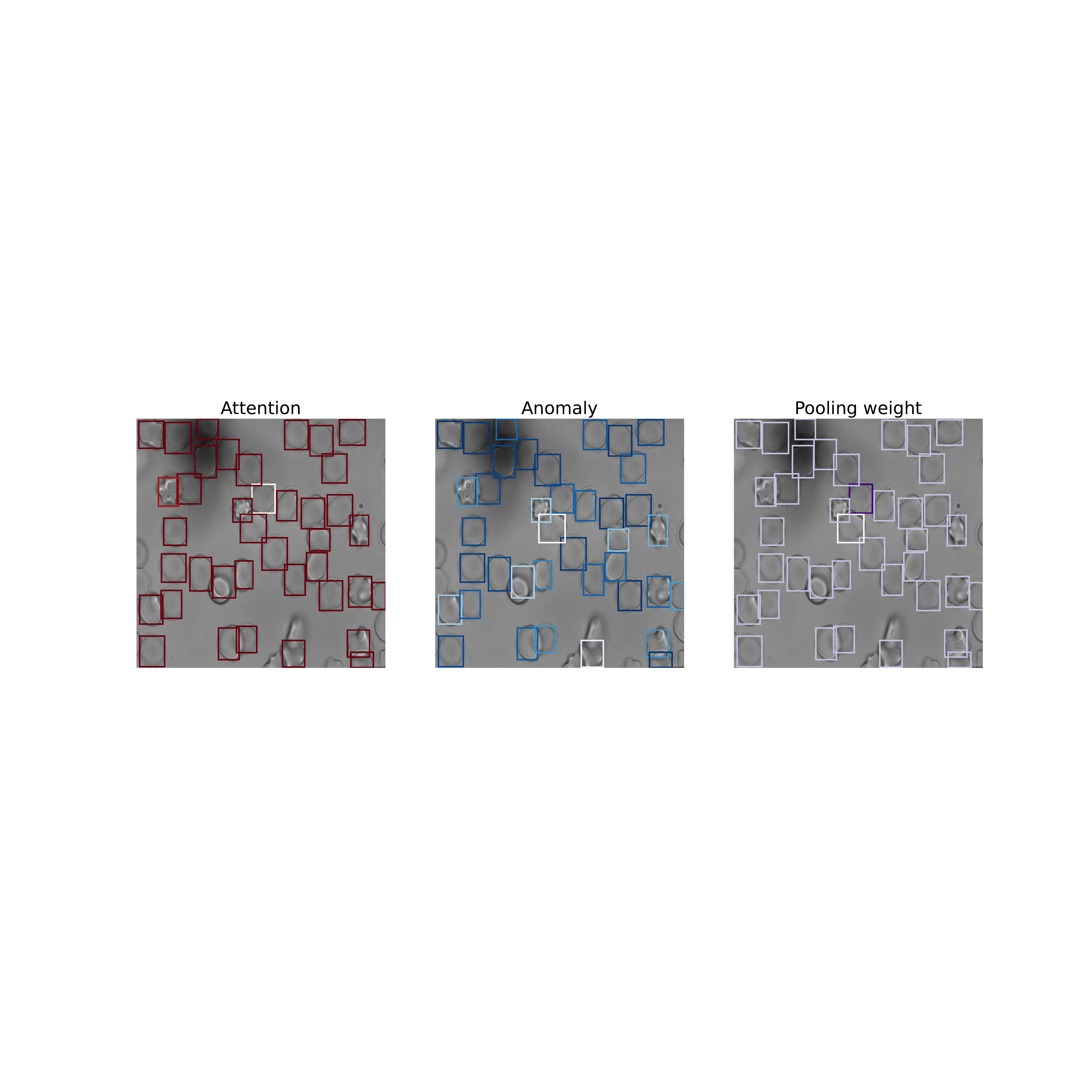}
		\caption*{Xero}
	\end{subfigure}
	\begin{subfigure}[b]{\w\textwidth}
		\centering
		\includegraphics[width=\textwidth]{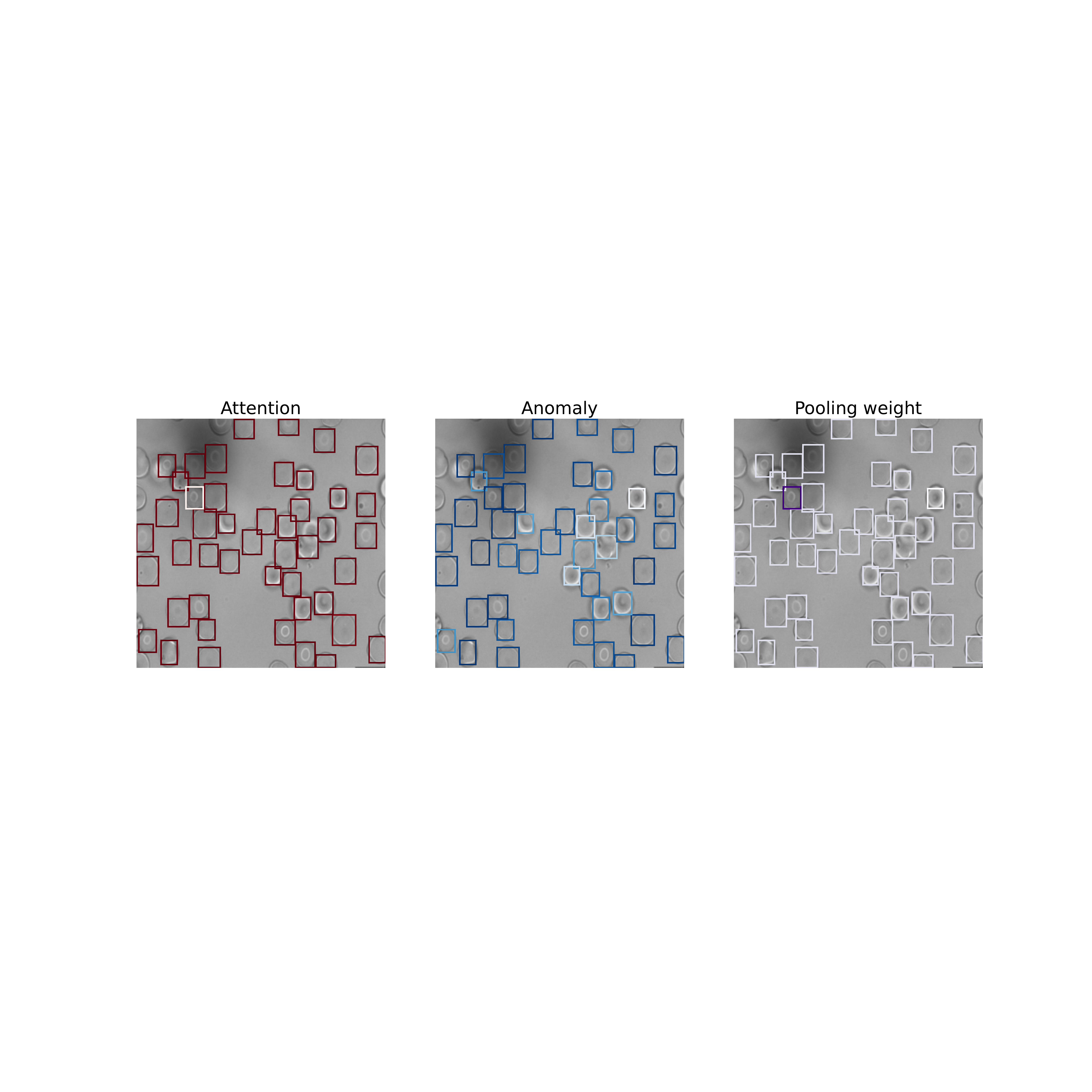}
		\caption*{HS}
	\end{subfigure}
	\begin{subfigure}[b]{\wb\textwidth}
		\centering
		\includegraphics[width=\textwidth]{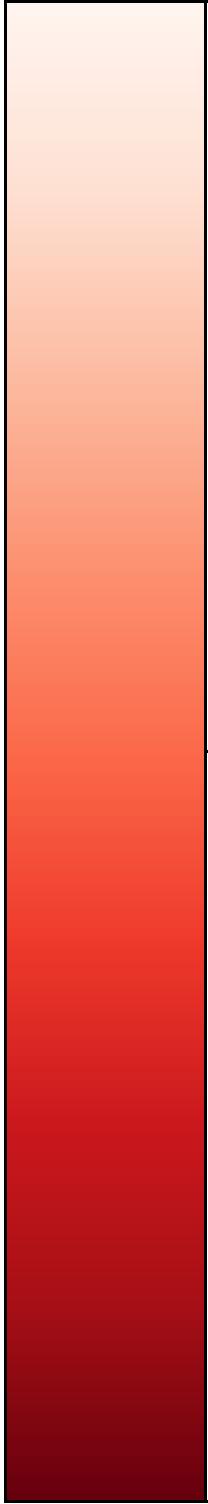}
		\caption*{ }
	\end{subfigure}
	\begin{subfigure}[b]{0.0142\textwidth}
		\centering
		\includegraphics[width=\textwidth]{images/arrow.pdf}
		\caption*{}
	\end{subfigure}
	\caption{Interpretation of bag classification at the instance level based on anomaly (top) and attention (bottom) scores. Compared with attention mechanism, disorder-relevant cells are more accurately scored by the anomaly mechanism. }
	\label{fig:interpretation}
\end{figure}

\subsection{Anomaly recognition}
\begin{figure}[t]
	\centering
	\includegraphics[width=\textwidth]{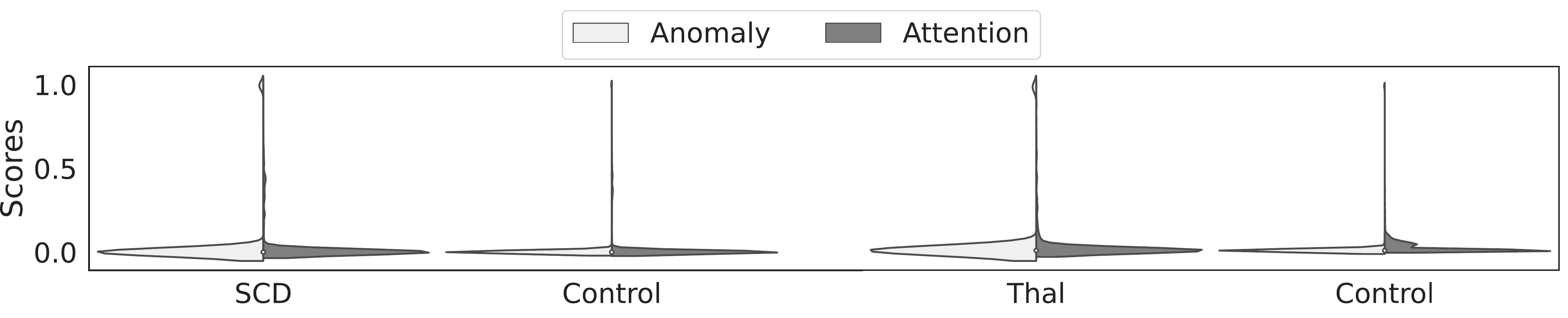}
	\caption{Distribution of anomaly scores compared to attention scores for SCD and Thal classes when their samples were absent during the training. Our method still can assign high anomaly scores to abnormal cells.}
	\label{fig:Anomaly}
	\vspace{-10pt}
\end{figure}

When our model estimates the distribution of control cells, high anomaly scores should be able to identify anomalous cells, whether the training set contains them or not.
This property of our method is useful for detecting anomalies in rare or unseen classes during training.
To examine our method in this context, we trained the model with the two HHA classes SCD and Thalassemia, separately.
Then, we tested the model with unseen class samples and measured their anomaly and attention scores.
Fig. \ref{fig:Anomaly} shows the results, where the small distribution of anomalous cells is clearly indicating anomaly awareness of the model. 
For both experiments, the distribution of scores in the control class is also shown.

\subsection{CO2 Emission Related to Experiments}
Experiments were conducted using our in-house infrastructure with a carbon efficiency of 0.432 kgCO$_2$eq/kWh. 
A cumulative 15 hours of computation was performed on the Tesla V100-SXM2-32GB (TDP of 300W) hardware.
Total emissions are estimated to be 1.94 kg CO$_2$eq.
Estimations were conducted using the \href{https://mlco2.github.io/impact#compute}{MachineLearning Impact calculator} presented in~\cite{lacoste2019quantifying}.

\section{Conclusion}
We introduce anomaly-aware MIL, a simple method for enhancing classification of hereditary hemolytic anemias.
Evaluations show that our method outperforms other MIL classifiers and provides an instance-level explanation for bag-level classification results. 
Moreover, our novel scoring mechanism identifies anomalies in rare disease, even if they have not been seen in training sets.
An interesting future work would be to upgrade the anomaly scoring to deep-learning-based out-of-distribution detection method.
Another interesting future work would be uncertainty estimation on anomalous instances, which might lead to a better understanding of models coverage in representation space.

\subsubsection{Acknowledgements}
The Helmholtz Association supports the present contribution under the joint research school “Munich School for Data Science - MUDS”.
C.M. has received funding from the European Research Council (ERC) under the European Union’s Horizon 2020 research and innovation programme (Grant agreement No. 866411).
CoMMiTMenT study was funded by the the European Seventh Framework Program under grant agreement number 602121 (CoMMiTMenT) and from the European Union’s Horizon 2020 Research and Innovation Programme. 
MemSID (NCT02615847) clinical trial was funded by the Foundation for Clinical Research Hematology for supporting the clinical trail at the Division of Hematology, University Hospital Zurich, and, partially, by the following foundations: Baugarten Zürich Genossenschaft und Stiftung, the Ernst Goehner Stiftung, the René und Susanna Braginsky Stiftung, the Stiftung Symphasis and the Botnar Foundation.”
Further funding for analysis of the obtained data was obtained European Union’s Horizon 2020 Research and Innovation Programme under grant agreement number 675115-RELEVANCE-H2020-MSCA-ITN-2015/H2020-MSCA-ITN-2015.

\bibliographystyle{splncs04}
\bibliography{mybibliography}

\begin{thebibliography}{10}
\providecommand{\url}[1]{\texttt{#1}}
\providecommand{\urlprefix}{URL }
\providecommand{\doi}[1]{https://doi.org/#1}

\bibitem{bessis2012corpuscles}
Bessis, M.: Corpuscles: Atlas of Red Blood Cell Shape. Springer Science \&
  Business Media (2012)

\bibitem{bi2021local}
Bi, Q., Yu, S., Ji, W., Bian, C., Gong, L., Liu, H., Ma, K., Zheng, Y.:
  Local-global dual perception based deep multiple instance learning for
  retinal disease classification. In: International Conference on Medical Image
  Computing and Computer-Assisted Intervention. pp. 55--64. Springer (2021)

\bibitem{campanella2019clinical}
Campanella, G., Hanna, M.G., Geneslaw, L., Miraflor, A., Werneck Krauss~Silva,
  V., Busam, K.J., Brogi, E., Reuter, V.E., Klimstra, D.S., Fuchs, T.J.:
  Clinical-grade computational pathology using weakly supervised deep learning
  on whole slide images. Nature medicine  \textbf{25}(8),  1301--1309 (2019)

\bibitem{fermo2021screening}
Fermo, E., Vercellati, C., Bianchi, P.: Screening tools for hereditary
  hemolytic anemia: new concepts and strategies. Expert review of hematology
  \textbf{14}(3),  281--292 (2021)

\bibitem{fujita2020cell}
Fujita, S., Han, X.H.: Cell detection and segmentation in microscopy images
  with improved mask r-cnn. In: Proceedings of the Asian Conference on Computer
  Vision (2020)

\bibitem{he2017mask}
He, K., Gkioxari, G., Doll{\'a}r, P., Girshick, R.: Mask r-cnn. In: Proceedings
  of the IEEE international conference on computer vision. pp. 2961--2969
  (2017)

\bibitem{he2016deep}
He, K., Zhang, X., Ren, S., Sun, J.: Deep residual learning for image
  recognition. In: Proceedings of the IEEE conference on computer vision and
  pattern recognition. pp. 770--778 (2016)

\bibitem{huisjes2018digital}
Huisjes, R., van Solinge, W., Levin, M., van Wijk, R., Riedl, J.: Digital
  microscopy as a screening tool for the diagnosis of hereditary hemolytic
  anemia. International Journal of Laboratory Hematology  \textbf{40}(2),
  159--168 (2018)

\bibitem{huisjes2020density}
Huisjes, R., Makhro, A., Llaudet-Planas, E., Hertz, L., Petkova-Kirova, P.,
  Verhagen, L.P., Pignatelli, S., Rab, M.A., Schiffelers, R.M., Seiler, E.,
  et~al.: Density, heterogeneity and deformability of red cells as markers of
  clinical severity in hereditary spherocytosis. haematologica
  \textbf{105}(2), ~338 (2020)

\bibitem{ilse2018attention}
Ilse, M., Tomczak, J., Welling, M.: Attention-based deep multiple instance
  learning. In: International conference on machine learning. pp. 2127--2136.
  PMLR (2018)

\bibitem{lacoste2019quantifying}
Lacoste, A., Luccioni, A., Schmidt, V., Dandres, T.: Quantifying the carbon
  emissions of machine learning. arXiv preprint arXiv:1910.09700  (2019)

\bibitem{li2019multi}
Li, S., Liu, Y., Sui, X., Chen, C., Tjio, G., Ting, D.S.W., Goh, R.S.M.:
  Multi-instance multi-scale cnn for medical image classification. In:
  International Conference on Medical Image Computing and Computer-Assisted
  Intervention. pp. 531--539. Springer (2019)

\bibitem{lu2021ai}
Lu, M.Y., Chen, T.Y., Williamson, D.F., Zhao, M., Shady, M., Lipkova, J.,
  Mahmood, F.: Ai-based pathology predicts origins for cancers of unknown
  primary. Nature  \textbf{594}(7861),  106--110 (2021)

\bibitem{lu2021data}
Lu, M.Y., Williamson, D.F., Chen, T.Y., Chen, R.J., Barbieri, M., Mahmood, F.:
  Data-efficient and weakly supervised computational pathology on whole-slide
  images. Nature biomedical engineering  \textbf{5}(6),  555--570 (2021)

\bibitem{reddi2019convergence}
Reddi, S.J., Kale, S., Kumar, S.: On the convergence of adam and beyond. arXiv
  preprint arXiv:1904.09237  (2019)

\bibitem{sadafi2020attention}
Sadafi, A., Makhro, A., Bogdanova, A., Navab, N., Peng, T., Albarqouni, S.,
  Marr, C.: Attention based multiple instance learning for classification of
  blood cell disorders. In: International Conference on Medical Image Computing
  and Computer-Assisted Intervention. pp. 246--256. Springer (2020)

\bibitem{sadafi2021sickle}
Sadafi, A., Makhro, A., Livshits, L., Navab, N., Bogdanova, A., Albarqouni, S.,
  Marr, C.: Sickle cell disease severity prediction from percoll gradient
  images using graph convolutional networks. In: Domain Adaptation and
  Representation Transfer, and Affordable Healthcare and AI for Resource
  Diverse Global Health, pp. 216--225. Springer (2021)

\bibitem{shi2020loss}
Shi, X., Xing, F., Xie, Y., Zhang, Z., Cui, L., Yang, L.: Loss-based attention
  for deep multiple instance learning. In: Proceedings of the AAAI Conference
  on Artificial Intelligence. vol.~34, pp. 5742--5749 (2020)

\bibitem{wu2021combining}
Wu, Y., Schmidt, A., Hern{\'a}ndez-S{\'a}nchez, E., Molina, R., Katsaggelos,
  A.K.: Combining attention-based multiple instance learning and gaussian
  processes for ct hemorrhage detection. In: International Conference on
  Medical Image Computing and Computer-Assisted Intervention. pp. 582--591.
  Springer (2021)

\end{thebibliography}

\end{document}